# CL-XABSA: Contrastive Learning for Cross-lingual Aspect-based Sentiment Analysis

*Nankai Lin, Yingwen Fu, Xiaotian Lin, Aimin Yang and Shengyi Jiang*

*Abstract*—As an extensive research in the field of natural language processing (NLP), aspect-based sentiment analysis (ABSA) is the task of predicting the sentiment expressed in a text relative to the corresponding aspect. Unfortunately, most languages lack sufficient annotation resources, thus more and more recent researchers focus on cross-lingual aspect-based sentiment analysis (XABSA). However, most recent researches only concentrate on cross-lingual data alignment instead of model alignment. To this end, we propose a novel framework, CL-XABSA: Contrastive Learning for Cross-lingual Aspect-Based Sentiment Analysis. Based on contrastive learning, we close the distance between samples with the same label in different semantic spaces, thus achieving a convergence of semantic spaces of different languages. Specifically, we design two contrastive strategies, token level contrastive learning of token embeddings (TL-CTE) and sentiment level contrastive learning of token embeddings (SL-CTE), to regularize the semantic space of source and target language to be more uniform. Since our framework can receive datasets in multiple languages during training, our framework can be adapted not only for XABSA task but also for multilingual aspect-based sentiment analysis (MABSA). To further improve the performance of our model, we perform knowledge distillation technology leveraging data from unlabeled target language. In the distillation XABSA task, we further explore the comparative effectiveness of different data (source dataset, translated dataset, and code-switched dataset). The results demonstrate that the proposed method has a certain improvement in the three tasks of XABSA, distillation XABSA and MABSA. For reproducibility, our code for this paper is available at https://github.com/GKLMIP/CL-XABSA.

*Index Terms*—Contrastive Learning, Cross-lingual Aspect-based Sentiment Analysis, Knowledge Distillation, Multilingual Aspect-based Sentiment Analysis

## I. INTRODUCTION

Aspect-based Sentiment Analysis (ABSA) is the task of predicting the sentiment expressed in a text relative to the corresponding aspect. Nowadays, ABSA task has gained a lot of attention [8][23][28][32]. Research on ABSA mainly focuses on high-resource languages, and there is limited research on low-resource languages. However, in real-world scenarios such as the E-commerce website, users' opinions are usually expressed in multiple different languages, including some low-resource ones [18][26].

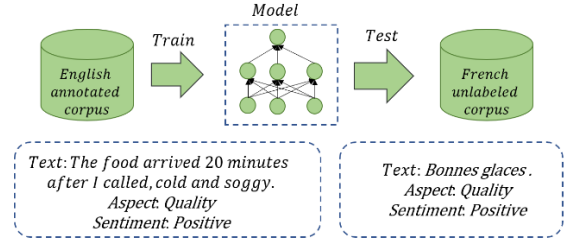

**Fig. 1.** An example of XABSA.

To tackle this issue, there has been an increasing interest in cross-lingual aspect-based sentiment analysis (XABSA) recently. In XABSA, only the annotated dataset of the source language $S$ is used for model training, and then the model is used to predict the dataset of the target language $T$. An example is shown in Figure 1 where we regard English as the source language and French as the target language. After training on the English annotated dataset, the model can directly identify the French sentence "Bonnes glaces." mentioned the aspect term "glaces", whose corresponding sentiment is positive.

Since there are few publicly available resources for low-resource languages, so scholars gradually consider the zero-shot transfer for cross-lingual aspect-based sentiment analysis task [16][30]. As far as we know, existing methods mainly implement the zero-shot method from the cross-lingual data alignment method. The data alignment method enables the model to indirectly learn the knowledge of the target language by means of data generation when there is no target language labeled dataset. Li et al. generated translation data in the target language by translating the labeled data into the source language [22]. Zhang et al. cleverly generated code-switched data suitable for XABSA tasks [37]. Their methods alleviate the XABSA data-hungry problem to a certain extent, but the implemented models are both BERT-based token classification models, which do not consider the zero-shot problem from the perspective of model alignment. The data alignment method simply merges the generated data with the source language data for training, while the model alignment can use the model to

Co-corresponding author: Aimin Yang and Shengyi Jiang
Nankai Lin, Yingwen Fu and Xiaotian Lin are the co-first authors.
Nankai Lin is with the School of Computer Science and Technology, Guangdong University of Technology, Guangzhou, Guangdong, PR China.
Yingwen Fu is with the School of Information Science and Technology, Guangdong University of Foreign Studies, Guangzhou, Guangdong, PR China.
Xiaotian Lin is with the School of Information Science and Technology, Guangdong University of Foreign Studies, Guangzhou, Guangdong, PR China.

Aimin Yang is with the School of Computer Science and Technology, Guangdong University of Technology, Guangzhou, Guangdong, PR China (e-mail: amyang@gdut.edu.cn).
Shengyi Jiang is with the Guangzhou Key Laboratory of Multilingual Intelligent Processing, and the School of Information Science and Technology, Guangdong University of Foreign Studies, Guangzhou, Guangdong, PR China. (e-mail: jiangshengyi@163.com).

learn the "alignment pattern" between cross languages. To a certain extent, the model alignment method can perform better on cross-language tasks.

Besides, the emergence of contrastive learning work has advanced significant progress in self-supervised representation learning [3][13][25][36]. The common idea of these works is as follows: shorten the distance between anchor points and "positive" samples in the embedding space and widen the distance between anchor points and many "negative" samples. Among them, positive examples are usually data-augmented samples generated from anchor points, and negative examples are the other samples in the same mini-batch. Khosla et al. [19] first extended contrastive learning to fully-supervised settings, which is called supervised contrastive learning. Supervised contrastive learning uses label information to distinguish positive and negative samples of anchor points. Contrastive learning has not only been shown to have superior performance on natural language processing [10] but is also applicable to cross-lingual tasks [5][11]. In cross-language NLP tasks, contrastive learning has shown its potential to learn the discrimination of samples with different labels in the feature space. In cross-lingual settings, contrastive learning can put the distance of samples with the same label from different languages closer in semantic space, so the vector space of the two languages can be implicitly brought closer.

To this end, we propose a novel framework CL-XABSA for XABSA based on contrastive learning. Specifically, there are two parts in CL-XABSA: token classification module and contrastive learning module. CL-XABSA uses different datasets for training, including labeled source language dataset, translation dataset from source language to target language, and code-switched datasets. code-switched datasets are constructed by switching the aspect terms between the source and translated target sentences to construct two bilingual sentences. Token classification module is a BERT-based sequence labeling model. The contrastive learning contains two alternative strategies, token level contrastive learning of token embeddings (TL-CTE) and sentiment level contrastive learning of token embeddings (SL-CTE). The contrastive learning module fuses and compares different datasets to learn richer language-independent knowledge and cross-lingual ability. Furthermore, we extend our framework to multilingual ABSA. To further improve the performance of our model, we perform knowledge distillation technology based on the unlabeled data from the target language.

In summary, the contributions of this paper are:

(1) To the best of our knowledge, we are the first to apply contrastive learning to the XABSA task. Our method can make the semantic space of different languages more consistent.

(2) We propose TL-CTE and SL-CTE to shorten the distance between tokens of the same label or the same sentiment in the source dataset, translated dataset, and code-switched datasets.

(3) By comparing different datasets, we further explore the effectiveness of contrastive learning in multi-teacher model distillation.

(4) Furthermore, we also apply contrastive learning to MABSA and distillation MABSA.

(5) The results show that the proposed method has a certain improvement in the three tasks of XABSA, multi-teacher distillation ABSA and multilingual distillation ABSA.

## II. RELATED WORK

### A. XABSA

Existing works on cross-lingual ABSA mainly focus on its sub-tasks including the cross-lingual aspect term extraction and aspect sentiment classification. The related works for XABSA are mainly drawn into two groups: data alignment and embedding learning.

Early methods for XABSA tasks mainly employ word alignment algorithms [9][30]. In addition, Zhou et al. [38] proposed to adopt a translate-then-align strategy to obtain pseudo-labeled data of the target language through machine translation. Zhang et al. cleverly generated code-switched data for XABSA tasks [37]. However, these methods heavily depend on the bilingual dictionary quality and the translation system performance.

In addition to data alignment, there are also some works investigating semantic representation construction to improve the XABSA performance. Ruder et al. [29] used cross-lingual word embeddings trained on a large parallel bilingual corpus for the XABSA task. By switching the word embeddings between different languages, the model can be used in a language-agnostic manner [2][1]. Jebbara and Cimiano incorporated multilingual word embeddings, which are based on a common shared vector space across various languages, into a convolutional neural network architecture for opinion target extraction [16].

### B. Contrastive Learning

Recently, contrastive learning has shown significant improvements for various NLP tasks, such as ABSA [17] and text classification [31]. Gao et al. [10] proposed a simple contrastive learning framework for semantic textual similarity (STS) tasks, which is adaptable to both unsupervised and supervised tasks. In addition, contrastive learning has received attention and applications in cross-lingual NLP tasks. Mohtarami et al. [24] introduced a novel memory network-based contrastive language adaptation approach to align the stances in the source and target languages exactly. Guo et al. proposed a contrastive learning method to obtain effective representations for cross-lingual text classification based on BERT model [11]. Choudhary et al. proposed to learn the representations of resource-poor and resource-rich language in a common emoji space based on contrastive learning [4].

### III. METHODOLOGY

In this section, we introduce the overall architecture of the proposed CL-XABSA. We regard the ABSA task as a sequence labeling task [12][21]. Give an input sentence S which is composed of a sequence of tokens $\{w_1, w_2, w_3, ..., w_n\}$ and a sequence of labels $\{l_1, l_2, l_3, ..., l_n\}$, where $n$ is the length of the sentence, the proposed model is to infer the label $l_i$ for each token $w_i$ and output a label sequence, where $l_i \in Y = \{B, I, E, S\} - \{POS, NEU, NEG\} \cup \{O\}$. $B, I, E$ respectively represent the beginning, middle, and ending of an aspect item and $S$ means that the aspect item is only one word. $POS, NEU$ and $NEG$ indicate that the sentiment of the corresponding aspect item is positive, neutral, and negative respectively. For example, $l_i = B - POS$ means $w_i$ is the beginning of a positive aspect term.

As shown in Figure 2. CL-XABSA is composed of two modules: 1) token classification module and 2) contrastive learning module. In the token classification module, the model regards ABSA as a sequence labeling task and chooses the label with the highest probability for each token. In contrastive learning module, we design two contrastive strategies, TL-CTE and SL-CTE.

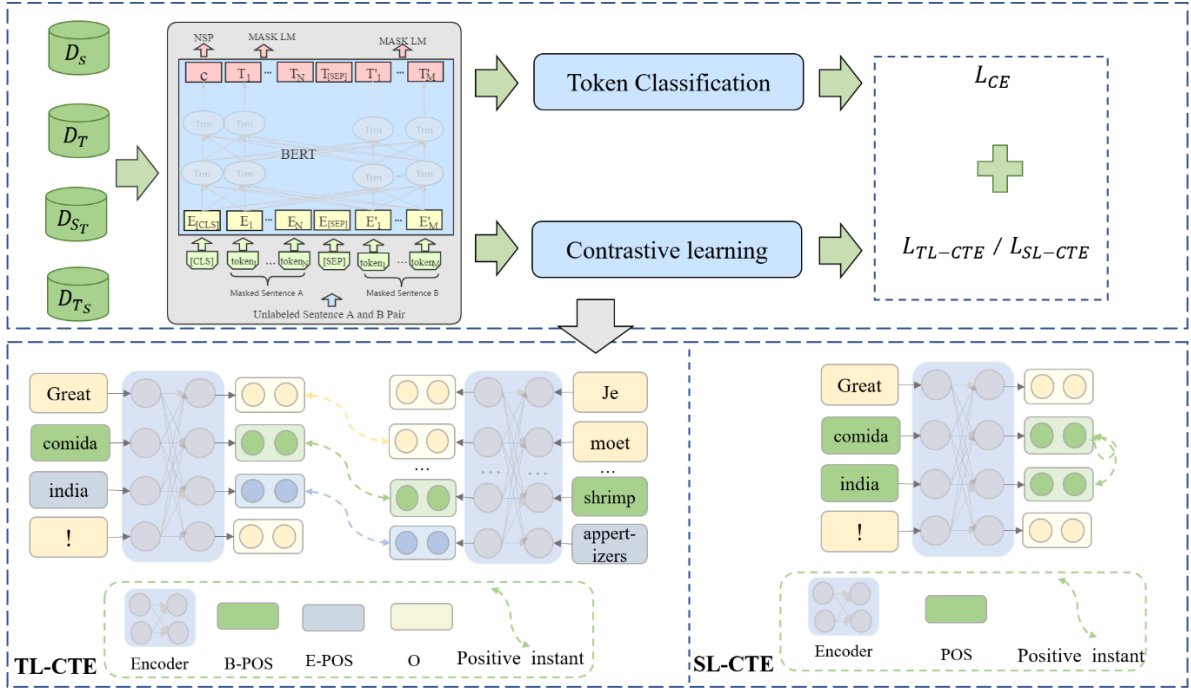

Fig. 2. The CL-XABSA framework.

Contrastive learning has been widely used in related research on natural language processing. Contrastive learning focuses on improving the model's ability to distinguish a given data point from "positive" examples (points that share the same label) and "negative" examples (different labels). The definition of "positive" examples determines which samples should be narrowed by the contrastive learning method. By defining the set of positive samples and set of negatives samples, a contrastive loss brings the latent representations of samples belonging to the same class closer together. Based on contrastive learning, we close the distance between samples with the same label in different semantic spaces, thus achieving a convergence of semantic spaces of different languages.

We propose TL-CTE and SL-CTE, which use the contrastive loss for XABSA tasks by defining positive examples from different levels. TL-CTE treats tokens with the same label as positive examples. On the other hand, SL-CTE performs contrastive learning from the sentiment level and treats tokens with the same sentiment as positive samples. The two strategies narrow the distance between tokens from different levels respectively. TL-CTE narrows the distance between tokens of the same label and SL-CTE compares tokens of the same sentiment. Therefore, TL-CTE is more fine-grained than SL-CTE. Through contrastive learning strategies at different levels, we can integrate the semantic spaces of different languages from different perspectives. Li et al. [39] proposed to use probability contrastive to replace feature contrastive, so that each module of the model can be updated instead of only the parameters of the representation layer. Therefore, in the contrastive learning module, we choose the method of probability contrastive.

In addition to leveraging labeled dataset in the source language $D_S$, we translate the source language $D_S$ to the target language and construct a translated dataset $D_T$. What's more, we utilize the code-switched datasets proposed by Zhang et al [37], which are called $D_{S_T}$ and $D_{T_S}$. $D_{S_T}$ is constructed by replacing the aspect terms appearing in the source language dataset $D_S$ with that one appearing in the target language dataset $D_T$. On the contrary, $D_{T_S}$ is generated by replacing the aspect terms appearing in the target language dataset $D_T$ with that one appearing in the source dataset $D_S$. Figure 3 shows an example of the source language dataset, translated dataset and code-switched dataset.

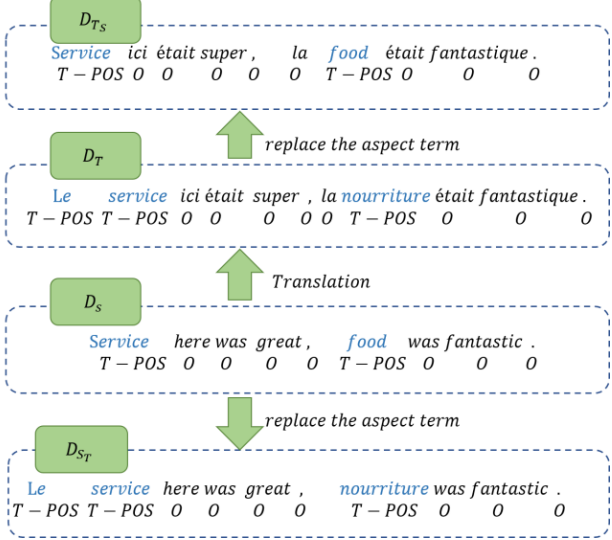

**Fig. 3.** An example of the code-switched dataset.

### A. Token Classification

We build the token classification model based on the BERT model. Given a text sequence $S = \{s_1, s_2, s_3, \ldots, s_n\}$ with $n$ words, the BERT model would encode $S$ into context-aware feature representations $h = \{h_1, h_2, h_3, \ldots, h_n\}$. Specifically, $h_i$ is represented as follows:

$$h_i = mBERT(s_i) \qquad (1)$$

Then, a multi-layer perceptron (MLP) classifier with the softmax function is leveraged to calculate the label probability distribution of $w_i$:

$$g_i = softmax(W \cdot h_i + b) \qquad (2)$$

where $W$ and $b$ are learnable parameters, and $g_i$ is the predicted label distribution of the $i$-th context-aware feature representation. Then the training objective $L_{CE}$ is computed as the cross-entropy loss:

$$L_{CE} = \frac{1}{N} \sum_0^N [-\frac{1}{n} \sum_{i=0}^n y_i log\,(g_i)] \qquad (3)$$

where $N$ refers to the number of sentences in the training set, $n$ is the number of tokens for each sentence, and $y_i$ represents the label of the $i$-th token.

### B. Contrastive Learning

**TL-CTE.** In order to narrow the distance between tokens of the same label, we propose TL-CTE to fuse and compare different datasets to learn richer language-independent knowledge and cross-lingual ability. As shown in Figure 4, we can see how TL-CTE applies contrastive learning on the

XABSA task. Specially, we denote a batch of sample and label pairs as $\{x_i, y_i\}_{i \in I}$, where $I = \{1, \cdots, K\}$ is the indices of the samples and $K$ is the batch size. Given a set of samples $P$ containing the same labels with the anchor, its positive set is defined by $P = \{p: p \in I, y_p = y_i \land p \neq i\}$, with size $|P|$, where $y_i \in Y_{token} = \{B, I, E, S\} - \{POS, NEU, NEG\} \cup \{O\}$ in TL-CTE. Our contrastive loss function for each entry $i$ across the batch is:

$$L_{TL_i} = \sum_{p \in P} log \frac{\exp(sim(g_i, g_p)/\tau)}{\sum_{k \in I/i} \exp(sim(g_i, g_k)/\tau)} \qquad (4)$$

where $sim(\cdot)$ indicates the cosine similarity function. The contrastive loss for the TL-CTE module is:

$$L_{TL-CTE} = \sum_{i=1}^K -\frac{1}{|P|} L_{TL_i} \qquad (5)$$

where $\tau$ is the temperature hyper-parameter. Larger values of $\tau$ scale down the dot-products, creating more difficult comparisons. After obtaining the contrastive loss through the TL-CTE module from Equation 5, we further weighted the contrastive loss $L_{TL-CTE}$ and the cross-entropy loss $L_{CE}$ to obtain the final model loss $L$:

$$L = \alpha * L_{TL-CTE} + (1 - \alpha) * L_{CE} \qquad (6)$$

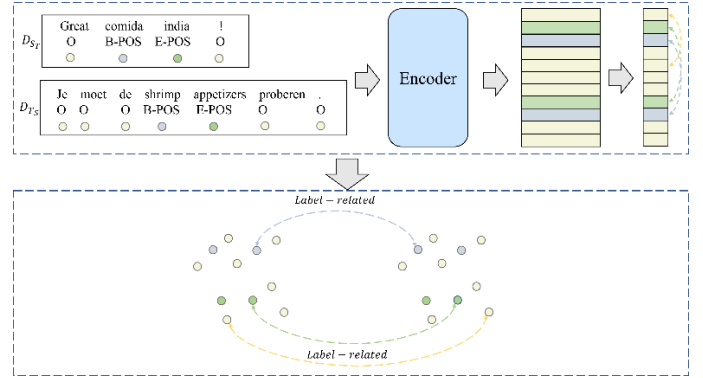

**Fig. 4.** The architecture of TL-CTE module.

**SL-CTE.** Different from TL-CTE, SL-CTE narrows the distance of samples at the sentiment level, which is shown in Figure 5. The positive set is given by $P = \{p: p \in I, y_p = y_i \land p \neq i\}$, with size $|P|$, where $y_i \in Y_{sen} = \{POS, NEU, NEG\} \cup \{O\}$ in SL-CTE. Besides, our contrastive loss function for each entry i across the batch is:

$$L_{SL_i} = \sum_{p \in P} log \frac{\exp(sim(g_i, g_p)/\tau)}{\sum_{k \in I/i} \exp(sim(g_i, g_k)/\tau)} \qquad (7)$$

Similar to TL-CTE module, the contrastive loss for the SL-CTE module is:

$$L_{SL-CTE} = \sum_{i=1}^{K} -\frac{1}{|P|} L_{SL_i} \qquad (8)$$

The final loss function $L$ also consists of two parts, cross entropy loss $L_{CE}$ and contrastive loss $L_{SL-CTE}$:

$$L = \alpha * L_{SL-CTE} + (1 - \alpha) * L_{CE} \qquad (9)$$

Although the computational process is similar to TL-CTE, SL-CTE performs contrastive learning from a coarser-grained sentence level, so its fitting goal is easier to achieve.

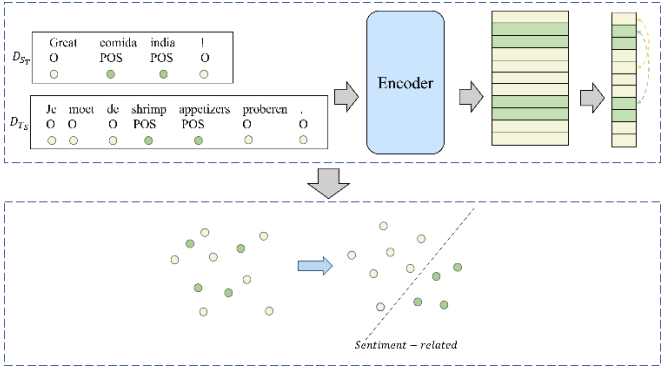

**Fig. 5.** The architecture of SL-CTE module.

### C. Distillation XABSA

Zhang et al. [37] verified the remarkable performance of knowledge distillation on the XABSA task, especially the multi-teacher distillation achieved the best performance on XABSA, therefore, we further applied our CL-XABSA framework to the multi-teacher distillation task.

Based on contrastive learning, we use translated dataset and code-switched datasets to train different teacher models, as shown in Figure 6. In order to fully exploit the knowledge in different datasets, we design a multi-teacher contrastive distillation model with different teacher models separately trained on three dataset combinations: $D_T \cup D_S$, $D_T \cup D_{S_T}$, $D_T \cup D_{T_S}$. Each teacher model contains the translated dataset $D_T$ to involve some language-specific knowledge and one of the remaining datasets to share the same sentence semantics ($D_S$), the same context sentence ($D_{T_S}$), and the same aspect term ($D_{S_T}$), respectively. Assuming that the prediction probability of the $k$-th teacher model is $g_{T_k}$, the label prediction probability obtained after fusing the three teacher models is

$$p_t = \sum_{k=1}^{3} \omega_k * g_{T_k} \qquad (10)$$

where $\omega_k$ is the weight for each teacher model. With the combined soft label $g_t$, a student model can be trained similarly by only using the token classification module. Since the student model receives more informative soft labels rather than hard labels [14], the student model replaces the original cross-entropy loss with mean squared error loss $L_{KD}$ during training:

$$L_{KD} = \frac{1}{|D_U|} \sum_{\vec{x} \in D_U} [\frac{1}{n} \sum_{j=0}^{n} MSE(g_{t_j}, g_{s_j})] \qquad (11)$$

Where $n$ is the number of tokens contained in a sentence and $j$ represents the $j$-th token in the sentence. In addition, $D_U$ denotes the unlabeled dataset in the target language and $g_{s_j}$ is the $j$-th token's predicted probability of the student model. We use the mean squared error loss (MSE($\cdot$)) to measure the difference between the two probability distributions.

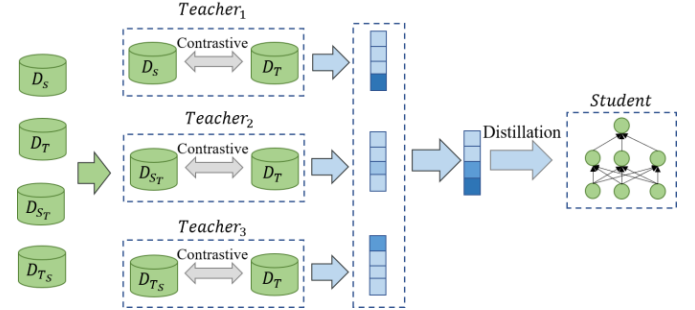

**Fig. 6.** The multi-teacher distillation process.

### D. Multilingual XABSA

We further adapt the CL-XABSA framework to make it suitable for the MABSA task. As shown in Figure 7, during the training phase, we utilize the source dataset $D_S$ for English language and multiple datasets for other four languages to train the teacher model. Specifically, considering the zero-shot setting, the multiple datasets contain target-language translated datasets $D_T$, code-switched datasets $D_{S_T}$ and $D_{T_S}$. What's more, we perform knowledge distillation technology leveraging data from unlabeled target language. The unlabeled data comes from the training set of the target languages provided by the original evaluation task, and then we discard its labels. In multilingual mode, whether it is a teacher model or a student model, the received data is the datasets containing multiple languages. Similar to the multi-teacher distillation task, the student model only uses the token classification module for learning and utilizes mean squared error loss as the loss function.

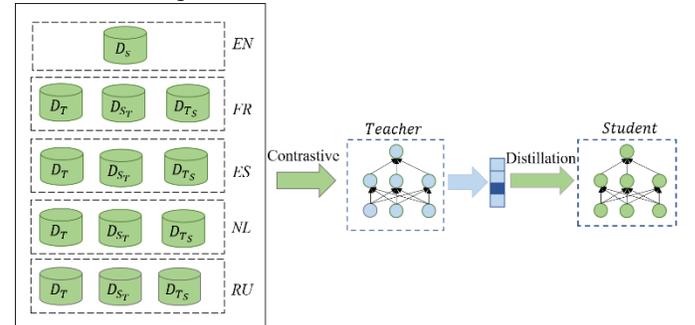

**Fig. 7.** The multilingual distillation process.

## IV. EXPERIMENTS

### A. Dataset

We conduct experiments on the SemEval-2016 dataset [27], which includes real user reviews in English (EN), French (FR), Spanish (ES), Dutch (NL), Russian (RU), and Turkish(TR).

However, due to the limitation of the Turkish data scale, many studies excluded this language during the experiment, similar to them, we also only used the other five languages during our experiments. We use the data processed by Zhang et al [37]. For each language, they not only split the data into training, validation, and testing sets, but also constructed code-switched datasets, where $D_{S_T}$ and the source language dataset $D_S$ have the same sentence context but aspect terms in different languages, while $D_{T_S}$ and the source language dataset $D_S$ have different sentence context but the same aspect term in the target language. The details of the dataset are shown in Table 1.

TABLE I
STATISTICS OF THE DATA IN EACH LANGUAGE.

| Dataset | Type | EN | FR | ES | NL | RU |
|---|---|---|---|---|---|---|
| Train | $D_S$ | 1600 | 1664 | 1656 | 1378 | 2924 |
| | $D_T$ | - | 1573 | 1571 | 1586 | 1577 |
| | $D_{S_T}$ | - | 1600 | 1600 | 1600 | 1600 |
| | $D_{T_S}$ | - | 1572 | 1571 | 1586 | 1577 |
| Dev | - | 400 | 332 | 414 | 344 | 731 |
| Test | - | 676 | 668 | 881 | 575 | 1209 |

B. Experimental Settings

We regard English as the source language and other languages as target languages. Similar to previous work [15][16][37], in order to simulate unsupervised setting, we use the English validation set in all experiments for the model selection. For the original training set of each target language, we discard the label of the training set and treat it as unlabeled data in the distillation stage [33][34][37]. For the training set with labels, we only use it in subsection 5.3 to verify the generalization of our MABSA model.

We conduct experiments based on the cased multilingual BERT (mBERT)[1] [7]. Similar to [37], we train the model up to 2000 steps and select the models on the last 500 steps. We initialize the student model with the TRANSLATION model (in subsection 4.1 Table 1 and subsection 4.2 Table 3), which is trained on the translated target dataset $D_T$, and then continue the training up to 1000 steps and select the models on the last 500 steps. When training the baseline methods ACS (in Table 1), TRANSLATION (in Table 1), MTL-AF (in Table 5), MTL-ACS (in Table 5), and MTL-ACS-D (in Table 5), we utilize a learning rate being 5e-5 and the batch size being 16, which are the same as the optimal parameters in [37]. For TL-CTE and SL-CTE, we try to train the model with a larger batch size, because a larger batch size is conducive to the training of contrastive learning. We select the best training hyper-parameters by conducting a grid search on batch size. The range of batch size is {16, 32, 64}. For TL-CTE, we use a batch size of 32 and for SL-CTE, we use the batch size being 64. It is worth mentioning that training the student model required more graphics memory, we could only set the batch size as 32 for SL-CTE. The results with different batch size are shown in Table 2. For the hyperparameter $\tau$, we set it as 0.07 consistently. For the multi-teacher distillation, we treat each teacher equally, which means $\omega_i = 1/3$ in Equation 10. The loss weighting factor $\alpha$ in Equation 6 and Equation 9 is set as 0.5.

Micro-F1 is employed as the evaluation metric where a prediction will be judged as correct only if both its boundary and sentiment polarity are correct. For all experiments, we report the average F1 scores over 5 runs with different random seeds.

V. RESULT AND ANALYSIS

A. Cross-lingual ABSA Results

We adopt the following approaches for comparisons: ZERO-SHOT, a method utilizing the labeled source data to fine-tune the model and directly conduct inference on the target data, which has shown to be a strong baseline for the cross-lingual adaptation [6][35]. To compare with the previous translation-based method, we adopt the baseline that utilizes the pseudo-labeled data with the Translate-then-Align paradigm [20][22] (TRANSLATION-TA) and the combination of the source data with such translated data (BILINGUAL-TA). In addition, we compared our method with the state-of-the-art method ACS [37].

TABLE II
THE RESULTS AMONG OUR METHODS AND OTHER COMPARATIVE METHODS. + DENOTES RESULTS ARE FROM LI ET AL. [22]

| Method | FR | ES | NL | RU | Avg. |
|---|---|---|---|---|---|
| ZERO-SHOT+ | 45.60 | 57.32 | 42.68 | 36.01 | 45.40 |
| TRANSLATION-TA+ | 40.76 | 50.74 | 47.13 | 41.67 | 45.08 |
| BILINGUAL-TA+ | 41.00 | 51.23 | 49.72 | 43.67 | 46.41 |
| TRANSLATION | 47.29 | 60.43 | 49.84 | 50.11 | 51.92 |
| ACS | 48.63 | 59.22 | 49.81 | 50.60 | 52.07 |
| CL-XABSA(TL) | 48.53 | 60.64 | **50.96** | **50.77** | 52.73 |
| CL-XABSA(SL) | **49.50** | **61.62** | 50.64 | 50.65 | **53.10** |

TABLE III
THE RESULTS OF DIFFERENT BATCH SIZES.

| Method | Batch Size | FR | ES | NL | RU | Avg. |
|---|---|---|---|---|---|---|
| TL-CTE | 16 | **49.33** | 60.12 | 49.61 | **51.51** | 52.64 |
| | 32 | 48.53 | **60.64** | **50.96** | 50.77 | **52.73** |
| | 64 | 49.02 | 59.85 | 50.62 | 50.24 | 52.43 |
| SL-CTE | 16 | 47.12 | 59.41 | 49.43 | **52.98** | 52.01 |
| | 32 | 49.08 | 60.25 | 50.13 | 50.62 | 52.52 |
| | 64 | **49.50** | **61.62** | 50.64 | 50.65 | **53.10** |

As shown in Table 2, both two contrastive strategies we proposed outperform the existing state-of-the-art models, where TL-CTE method performs better on French and Spanish, while on Dutch and Russian, SL-CTE method performs better. On average, the SL-based method outperforms the TL-based method on the XABSA task, with an average Micro-F1 value reaching 53.10%. As far as we know, the performance of the contrastive learning model is significantly affected by the batch size [3]. Therefore, we explore the optimal batch size for our two strategies, and the results are shown in Table 3. We can see that for the SL-CTE strategy, the larger the batch size, the better the performance the model can obtain, while for the TL-CTE strategy, the optimal batch size value is 32. Since in a batch, the more negative examples, the better the performance [19], and when the batch size is the same, for each sample in CL-XABSA(TL), the number of negative examples is higher than that in CL-XABSA(SL). That is, CL-XABSA(SL) needs to have a larger batch size to achieve the number of negative examples required for optimal performance.

*B. Multi-teacher Distillation Results*

In order to improve the performance of the model, we further introduce distillation technology. As the results shown in Table 4, it is obvious to see that multi-teacher distillation can effectively improve the XABSA performance. During multi-teacher distillation, although the result of CL-XABSA(TL) model is slightly lower than that of ACS model, our framework still shows a certain improvement that the performance of CL-XABSA(SL) model is 0.56 higher than that of ACS model. We believe that due to the limitation of the data size, the initialized student model and teacher model of CL-XABSA are difficult to get better model performance. Therefore, we further analyze the performance of initialized student model and different teacher models. The results are shown in Table 5. We can see from the table that when using $D_T$ to train the model, TL-CTE and SL-CTE only aim to shorten the distance between token and sentiment in the monolingual vector space, which is similar to the goal of the monolingual sentiment classification model, so the ACS model outperforms the contrastive learning models. As for the model trained by $D_S \cup D_T$ and $D_{S_T} \cup D_T$, the correlation between the source and target languages is relatively limited compared with the model trained by $D_{T_S} \cup D_T$. Specifically, aspect terms are language-independent in $D_S$ and $D_T$. On the contrary, $D_{S_T}$ and $D_T$ are only associated with aspect term information, which means that their semantic representations are not in a uniform space. However, as mentioned above, since the TL-CTE model needs to regularize more token spaces to be more uniform, its performance can only be better improved on the datasets that contain more information. As for the models trained on $D_{S_T} \cup D_T$ and $D_S \cup D_T$, the cross-language information provided by these two corpora is not enough to support the TL-CTE model to achieve good results, while the SL-CTE model that requires less learning information can achieve a better effect. In light of the dataset of $D_{T_S} \cup D_T$, aspect terms for different languages only appear in the semantic representation of the target language. Although the semantic space of the source language is lacking, the label information provided by the aspect term can better improve the XABSA task, which indicates that TL-CTE can perform better when there is enough information for learning.

TABLE IV
THE RESULTS OF MULTI-TEACHER DISTILLATION EXPERIMENTS.

| Method | FR | ES | NL | RU | Avg. |
|---|---|---|---|---|---|
| ACS + Multi-teacher Distillation | 50.13 | **63.82** | 53.51 | 54.18 | 55.41 |
| CL-XABSA(TL) + Multi-teacher Distillation | 51.57 | 63.51 | 52.87 | 53.08 | 55.26 |
| CL-XABSA(SL) + Multi-teacher Distillation | **51.69** | 63.54 | **54.28** | 53.98 | **55.97** |

TABLE V
THE RESULTS OF INITIALIZED STUDENT MODEL AND DIFFERENT TEACHER MODELS.

| Data | Method | FR | ES | NL | RU | Avg. |
|---|---|---|---|---|---|---|
| $D_T$ | ACS | **47.29** | 60.43 | **49.84** | **50.11** | **51.92** |
| | CL-XABSA(TL) | 46.96 | 60.20 | 49.42 | 49.39 | 51.49 |
| | CL-XABSA(SL) | 46.34 | 59.85 | 49.38 | 48.96 | 51.13 |
| $D_S \cup D_T$ | ACS | 47.34 | 60.06 | 50.61 | **51.56** | 52.39 |
| | CL-XABSA(TL) | 48.87 | 60.50 | 51.20 | 50.72 | 52.82 |
| | CL-XABSA(SL) | **49.84** | **61.29** | **51.22** | 50.39 | **53.19** |
| $D_{S_T} \cup D_T$ | ACS | 46.93 | 59.48 | 48.77 | **50.24** | 51.36 |
| | CL-XABSA(TL) | **47.56** | 59.12 | 48.68 | 50.14 | 51.38 |
| | CL-XABSA(SL) | 46.58 | **60.78** | **50.55** | 49.87 | **51.95** |
| $D_{T_S} \cup D_T$ | ACS | 46.87 | 59.15 | 50.43 | **49.52** | 51.49 |
| | CL-XABSA(TL) | **48.43** | **59.97** | **50.87** | 48.97 | **52.06** |
| | CL-XABSA(SL) | 47.15 | 59.20 | 49.41 | **49.58** | 51.34 |

TABLE VI
THE RESULTS OF MULTILINGUAL DISTILLATION EXPERIMENTS. + DENOTES RESULTS ARE FROM LI ET AL. [22]

| Method | FR | ES | NL | RU | Avg. |
|---|---|---|---|---|---|
| MTL-TA[+] | 40.72 | 54.14 | 49.06 | 43.89 | 46.95 |
| MTL-WS[+] | 46.93 | 58.18 | 49.87 | 44.88 | 49.96 |
| MTL-AF | 48.73 | 59.54 | 52.15 | 50.99 | 52.85 |
| MTL-ACS | 48.60 | 60.22 | 51.43 | 50.36 | 52.65 |
| MTL-CL-XABSA(TL) | **50.01** | 59.05 | 51.22 | 50.59 | **52.72** |
| MTL-CL-XABSA(SL) | 49.92 | 58.54 | 49.70 | 49.76 | 51.98 |
| MTL-ACS-D | 49.70 | 61.65 | 52.41 | 52.11 | 53.97 |
| MTL-CL-XABSA(TL)-D | **53.03** | 62.01 | 54.04 | **54.63** | **55.93** |
| MTL-CL-XABSA(SL)-D | 52.23 | **62.19** | **54.25** | 53.87 | 55.64 |

## C. Multilingual Distillation Results

In addition to XABSA, we further report on our work on MABSA. As shown in Table 6, after knowledge distillation, CL-XABSA based on TL-CTE achieves the best results with an average Micro-F1 reaching 55.93, which is 1.94 higher than the state-of-the-art model MTL-ACS-D. Furthermore, although the CL-XABSA model based on SL-CTE does not exceed the CL-XABSA based on TL-CTE, it still has a significant improvement with the average Micro-F1 value being 1.67 higher than the MMTL-ACS-D model. On the contrary, for the model without knowledge distillation, the teacher model of our method does not outperform the existing models (MTL-ACS and MTL-AF), while the student model achieves better performance.

TABLE VII
THE RESULTS OF MULTILINGUAL DISTILLATION
EXPERIMENTS IN TRAINING DATASETS.

| Method | Dataset | FR | ES | NL | RU | Avg. |
|---|---|---|---|---|---|---|
| MTL-ACS | Train | 48.60 | 60.22 | 51.43 | 50.36 | 53.13 |
| | Test | 53.57 | 58.70 | 49.23 | 52.90 | |
| CL-XABSA (TL) | Train | 50.01 | 59.05 | 51.22 | 50.59 | **53.80** |
| | Test | 55.84 | 58.34 | 50.99 | 54.32 | |
| CL-XABSA (SL) | Train | 49.92 | 58.54 | 49.70 | 49.76 | 53.43 |
| | Test | 55.84 | 58.34 | 50.99 | 54.32 | |

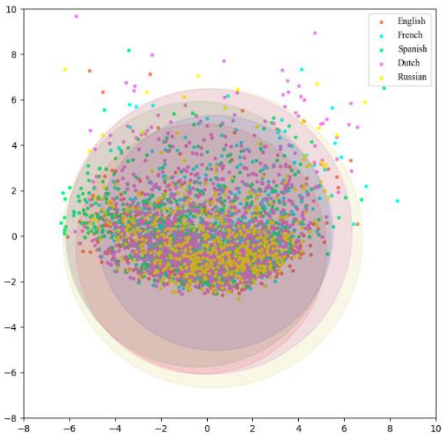

(a.1)

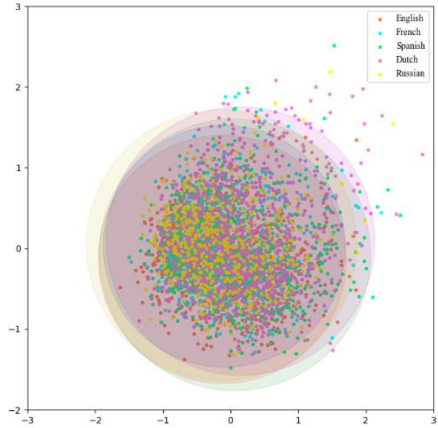

(b.1)

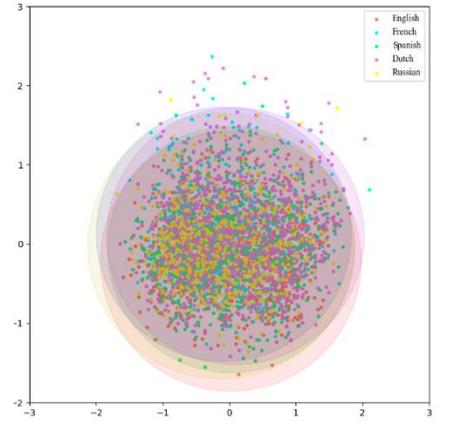

(c.1)

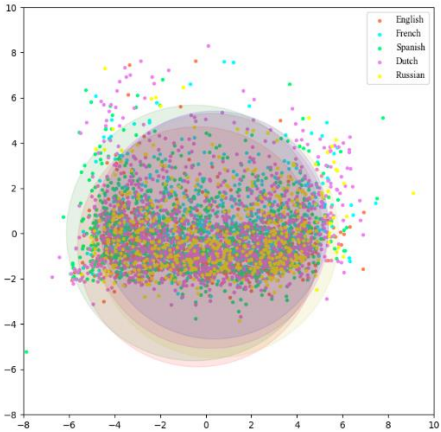

(a.2)

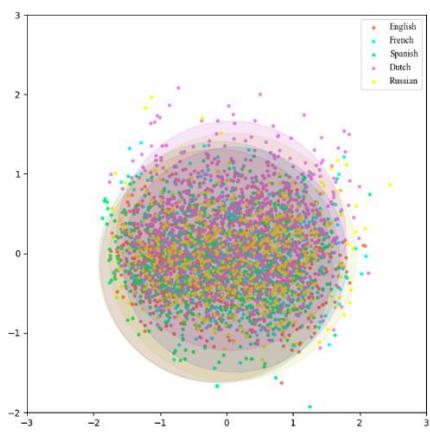

(b.2)

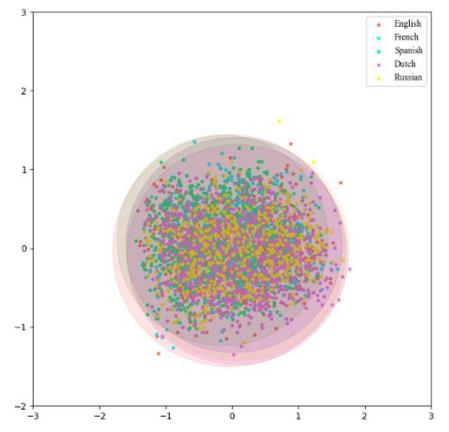

(c.2)

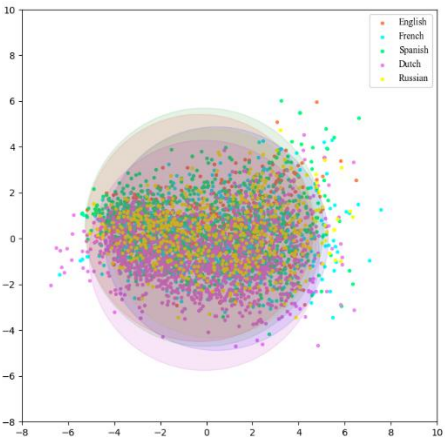

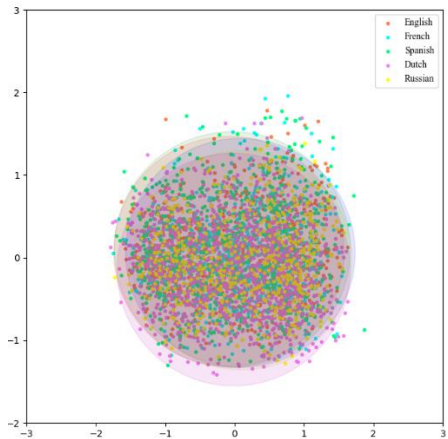

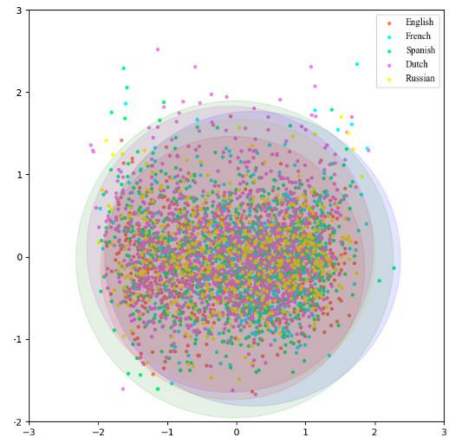

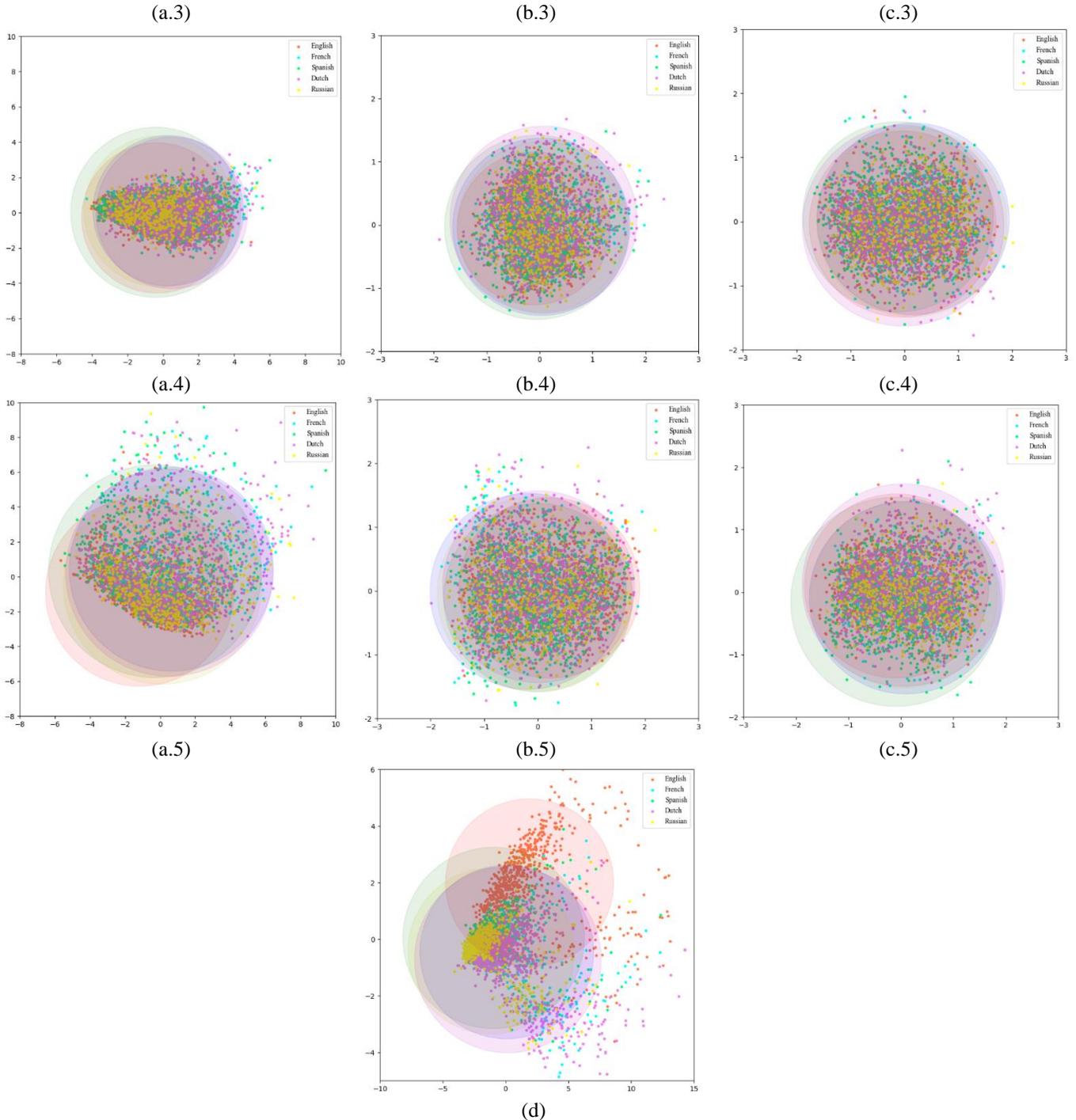

**Fig. 8.** 2D visualization of different models' semantic space. Subfigures (a.*) are the MTL-ACS. Subfigures (b.*) are the MTL-CL-XABSA(SL). Subfigures (c.*) are the MTL-CL-XABSA(TL). Subfigures (d) is the mbert.

To explore the reasons for this result, we further compared the performance of our model with the MTL-ACS model on the target language training set provided by the SemEval workshop (i.e. distillation data). The result is shown in Table 7. On distillation data, our model outperforms other models, and our (teacher) model performs better in terms of the average results of training and test sets, meaning that the training and test set distribution provided by the SemEval workshop are inconsistent and our model has strong generalization performance. Therefore, our model overall outperforms existing models on dataset in different distribution spaces.

### D. Semantic Space Verification

To demonstrate the ability of our approach in achieving a convergence of semantic spaces of different languages, we plot 2D visualization of the hidden sentence representations of different models using PCA. We visualized the results of MTL-ACS, MTL-CL-XABSA(TL), MTL-CL-XABSA(TL), and mBERT models without fine-tuning respectively. In addition to the other three methods except for mBERT, we visualize the results of five experiments. The experimental results are shown in Figure 8. We treat the samples with the same language as the same cluster. It can be seen that in the semantic space of mbert, the semantic space of each language is relatively scattered, while the other three methods will be closer. In additional, although the other three methods close the semantic space of different languages, the semantic space of MTL-ACS is more scattered than other two methods we proposed, which means that MTL-CL-XABSA (TL) and MTL-CL- XABSA(TL) better achieves convergence of semantic spaces of different languages.(the horizontal and vertical coordinates of the subgraphs (a.*) are relatively large).

In order to more intuitively measure the consistency of the semantic space, we use the classical clustering evaluation metric——Calinski-Harabasz metric for evaluation. When the Calinski-Harabasz metric is larger, the boundaries between different clusters are more obvious; when the Calinski-Harabasz metric is smaller, the degree of coincidence between different clusters is higher, that is, the semantic space representation is more consistent. As shown in Table 8, the mBERT model without fine-tuning has a large Calinski-Harabasz value. Although MTL-ACS can narrow the semantic space by data alignment to a certain extent, its Calinski-Harabasz value is higher than our two methods. From the Calinski-Harabasz metric point of view, MTL-CL-XABSA(TL) works best on the goal of achieving semantic space consistency.

TABLE VIII
THE CALINSKI-HARABASZ INDEX OF DIFFERENT MODELS.

| Model | Calinski-Harabasz index |
|---|---|
| mBERT | 264.10 |
| MTL-ACS | 37.40 |
| MTL-CL-XABSA(TL) | 22.16 |
| MTL-CL-XABSA(SL) | 23.09 |

V. CONCLUSION

To the best of our knowledge, we are the first to apply contrastive learning to XABSA and MABSA tasks. The results show the superior performance and generalization of our framework CL-XABSA. At the same time, we also verify the effectiveness of knowledge distillation on XABSA and MABSA tasks. Our framework can be applied to a variety of multilingual pre-trained models. Among the mbert-based models, our model achieves the best performance, and our method could be regarded as a new state-of-the-art for XABSA and MABSA tasks. In the future, we will further explore the possibility of contrastive learning on XABSA tasks, and explore contrastive learning methods to better learn cross-language alignment patterns.